\pdfoutput=1

\documentclass[11pt]{article}

\usepackage{naacl2021}

\usepackage{times}
\usepackage{latexsym}

\usepackage[T1]{fontenc}

\usepackage[utf8]{inputenc}

\usepackage{microtype}

\usepackage{times}
\usepackage{latexsym}
\usepackage[bottom]{footmisc}
\usepackage{graphicx}
\usepackage{booktabs} 
\usepackage{multirow}
\usepackage{marvosym}
\usepackage{arydshln}

\usepackage{microtype}

\usepackage{xcolor}
\usepackage{amsmath}

\newcommand{\loss}{\ensuremath\mathcal{L}}

\title{Universal Adversarial Attacks with Natural Triggers for Text Classification}

\author{Liwei Song$^\S$  \qquad Xinwei Yu$^\S$  \qquad Hsuan-Tung Peng$^\S$  \qquad Karthik Narasimhan \\
         Princeton University \\ 
         \texttt{\{liweis, xinweiy, hsuantungp,  karthikn\}@princeton.edu}}

\date{}

\begin{document}
\maketitle
\begingroup\renewcommand
\thefootnote{}
\footnotetext{$^\S$Equal contribution
}
\endgroup

\begin{abstract}
Recent work has demonstrated the vulnerability of modern text classifiers to \emph{universal adversarial attacks}, which are input-agnostic sequences of words added to text processed by classifiers.
Despite being successful, the word sequences produced in such attacks are often ungrammatical and can be easily distinguished from natural text. We develop adversarial attacks that appear closer to natural English phrases and yet confuse classification systems when added to benign inputs. We leverage an adversarially regularized autoencoder (ARAE)~\cite{zhao_ARAE_ICML18} to generate triggers and propose a gradient-based search that aims to maximize the downstream classifier's prediction loss. 
Our attacks effectively reduce model accuracy on classification tasks while being less identifiable than prior models as per automatic detection metrics and human-subject studies. Our aim is to demonstrate that adversarial attacks can be made harder to detect than previously thought and to enable the development of appropriate defenses.\footnote{Our code is available at \url{https://github.com/Hsuan-Tung/universal_attack_natural_trigger}.}

\end{abstract}

\section{Introduction}

Adversarial attacks have recently been quite successful in foiling neural text classifiers~\cite{jia_adv_NLP_EMNLP17, ebrahimi_adv_NLP_ACL18}.
\textit{Universal adversarial attacks}~\cite{Wallace_uni_adv_NLP_EMNLP19,behjati_uni_adv_NLP_ICASSP19} are a sub-class of these methods where the same attack perturbation can be applied to \emph{any input to the target classifier}. These attacks, being input-agnostic, point to more serious shortcomings in trained models since they do not require re-generation for each input. 
However, the attack sequences generated by these methods are often meaningless and irregular text (e.g., ``zoning tapping fiennes'' from \citet{Wallace_uni_adv_NLP_EMNLP19}). While human readers can easily identify them as unnatural, one can also use simple heuristics to spot such attacks. For instance, the words in the above attack trigger have an average frequency of $14$ compared to $6700$ for words in benign inputs in the Stanford Sentiment Treebank (SST)~\cite{Socher2013RecursiveDM}.

In this paper, we design \emph{natural attack triggers} by using an adversarially regularized autoencoder (ARAE)~\cite{zhao_ARAE_ICML18}, which consists of an auto-encoder and a generative adversarial network (GAN). We develop a \emph{gradient-based search} over the noise vector space to identify triggers with a good attack performance. Our method -- Natural Universal Trigger Search (NUTS) -- uses projected gradient descent with $l_2$ norm regularization to avoid using out-of-distribution noise vectors and maintain the naturalness of text generated.\footnote{We define naturalness in terms of how likely a human can detect abnormalities in the generated text.}

Our attacks perform quite well on two different classification tasks -- sentiment analysis and natural language inference (NLI). 
For instance, the phrase \textit{combined energy efficiency}, generated by our approach, results in a classification accuracy of 19.96\% on negative examples on the Stanford Sentiment Treebank~\cite{Socher2013RecursiveDM}.
Furthermore, we show that our attack text does better than prior approaches on three different measures -- average word frequency, loss under the GPT-2 language model~\cite{gpt2_openai_2019}, and errors identified by two online grammar checking tools~\cite{scribens, cheggwriting}. 
A human judgement study shows that up to 77\% of raters find our attacks more natural than the baseline and almost 44\% of humans
find our attack triggers concatenated with benign inputs to be natural.
This demonstrates that using techniques similar to ours, adversarial attacks could be made much harder to detect than previously thought and we require the development of appropriate defenses in the long term for securing our NLP models.

\section{Related Work}

\paragraph{Input-dependent attacks}
These attacks generate specific triggers for each different input to a classifier. \citet{jia_adv_NLP_EMNLP17} fool reading comprehension systems by adding a single distractor sentence to the input paragraph.
\citet{ebrahimi_adv_NLP_ACL18} replace words of benign texts with similar tokens using word embeddings.
Similarly, \citet{alzantot_adv_NLP_EMNLP18} leverage genetic algorithms to design word-replacing attacks.
\citet{zhao_adv_NLP_ICLR18} adversarially perturb latent embeddings and use a text generation model to perform attacks. 
\citet{song2020adversarial} develop natural attacks to cause semantic collisions, i.e. make texts that are semantically unrelated judged as similar by NLP models.

\paragraph{Universal attacks}
Universal attacks are input-agnostic and hence, word-replacing and embedding-perturbing approaches are not applicable. \citet{Wallace_uni_adv_NLP_EMNLP19} and \citet{behjati_uni_adv_NLP_ICASSP19} concurrently proposed to perform gradient-guided searches over the space of word embeddings to choose attack triggers.
In both cases, the attack triggers are meaningless and can be easily detected by a semantic checking process.
In contrast, we generate attack triggers that appear more natural and retain semantic meaning.  
In computer vision, GANs have been used to create universal attacks~\cite{xiao_GAN_attack_IJCAI18,poursaeed_GAN_attack_CVPR18}.
Concurrent to our work, \citet{atanasova2020generating} design label-consistent natural triggers to attack fact checking models. They first predict unigram triggers and then use a language model conditioned on the unigram to generate natural text as the final attack, while we generate the trigger directly.

\section{Universal Adversarial Attacks with Natural Triggers}

We build upon the universal adversarial attacks proposed by \citet{Wallace_uni_adv_NLP_EMNLP19}. To enable \textit{natural} attack triggers, we use a generative model which produces text using a continuous vector input, and perform a gradient-guided search over this input space. The resulting trigger, which is added to benign text inputs, is optimized so as to maximally increase the loss under the target classification model.

\paragraph{Problem formulation}
Consider a pre-trained text classifier $F$ to be attacked. Given a set of benign input sequences $\{x\}$ with the same ground truth label $y$, the classifier has been trained to predict $F(x) = y$.
Our goal is to find a single input-agnostic trigger, $t$, that when concatenated\footnote{We follow \citet{Wallace_uni_adv_NLP_EMNLP19} in adding the triggers in front of the benign text.} with \textit{any} benign input, causes $F$ to perform an incorrect classification, i.e., $F([t;x]) \neq y$, where $;$ represents concatenation. In addition, we also need to ensure the trigger $t$ is natural fluent text.

\begin{figure}[t]
	\centering
	\includegraphics[width=\linewidth]{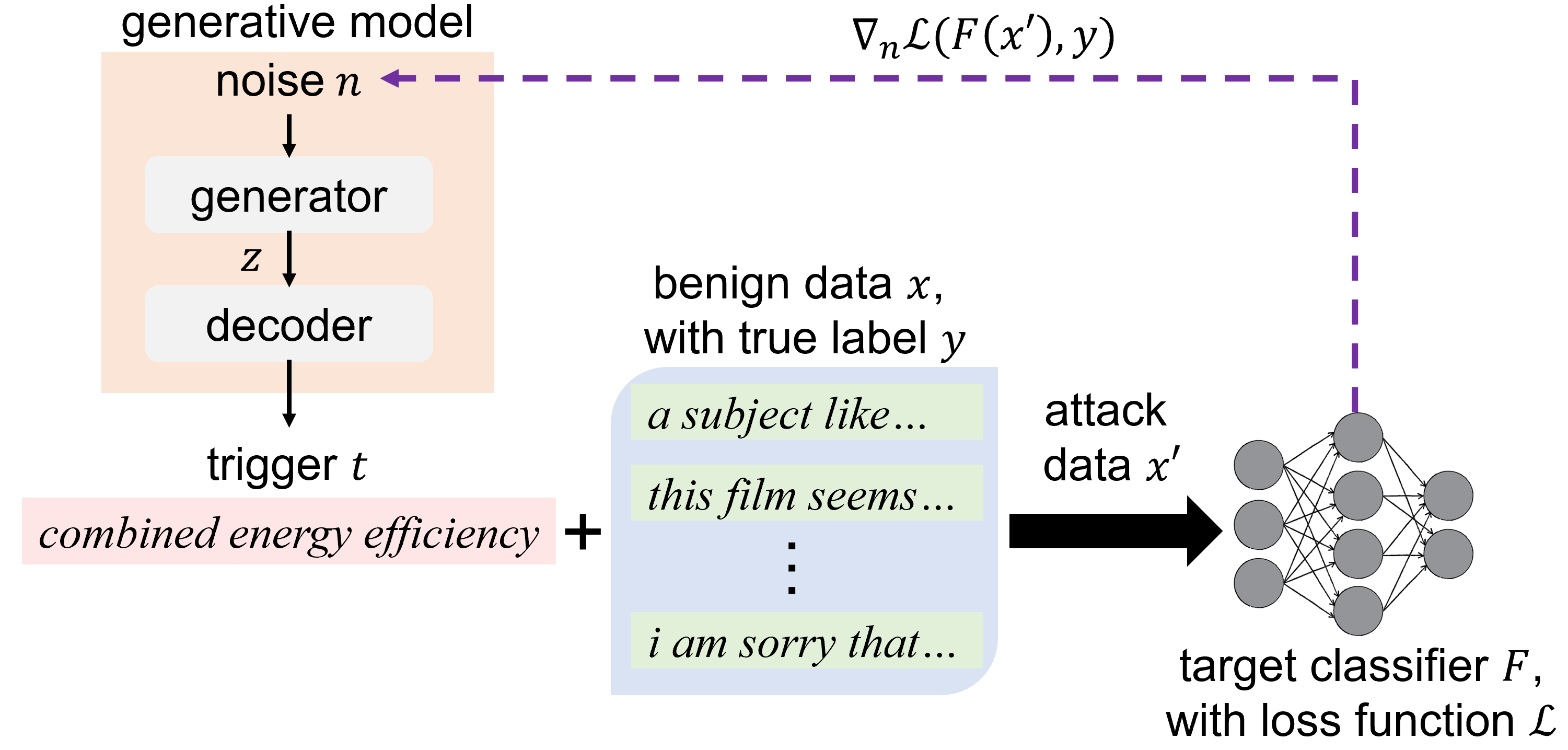}
	\caption{\textbf{Overview of our attack.} Based on the gradient of the target model's loss function, we iteratively update the noise vector $n$ with small perturbation to obtain successful and natural attack triggers.	}
	\label{fig:gradient_attack}
\end{figure}

\paragraph{Attack trigger generation}
To ensure the trigger is natural, fluent and carries semantic meaning, we use a pre-trained adversarially regularized autoencoder (ARAE)~\cite{zhao_ARAE_ICML18} (details in Section~\ref{sec:experiments}). The ARAE consists of an encoder-decoder structure and a GAN \cite{goodfellow_GAN_NIPS14}. 
The input is a standard Gaussian noise vector $n$, which is first mapped to a latent vector $z$ by the generator.  
Then the decoder uses this $z$ to generate a sequence of words -- in our case, the trigger $t$. 
This trigger is then concatenated with a set of benign texts $\{x\}$ to get full attack texts $\{x'\}$. The overall process can be formulated as follows:
\begin{equation*}
\begin{aligned}
z =  \textsc{Generator}(n) &;
t =  \textsc{Decoder}(z); \\
x' &=  [t;x]
\end{aligned}
\end{equation*}

We then pass each $x'$ into the target classifier and compute the gradient of the classifier's loss with respect to the noise vector, $\nabla_{n}\loss(F(x'), y)$. Backpropagating through the decoder is not straightforward since it produces discrete symbols. Hence, we use a reparameterization trick similar to the trick in Gumbel softmax~\cite{Jang_Gumbel_Softmax_ICLR17} to sample words from the output vocabulary of ARAE model as a one-hot encoding of triggers, while allowing gradient backpropagation.
Figure~\ref{fig:gradient_attack} provides an overview of our attack algorithm, which we call Natural Universal Trigger Search (NUTS).

\paragraph{Ensuring natural triggers}

In the ARAE model, the original noise vector $n_{0}$ is sampled from a standard multi-variant Gaussian distribution. While we can change this noise vector to produce different outputs, simple gradient search may veer significantly off-course and lead to bad generations. 
To prevent this, following \citet{carlini_adv_img_SP17}, we use projected gradient descent with an $l_2$ norm constraint to ensure the noise $n$ is always within a limited ball around $n_{0}$.
We iteratively update $n$ as:
\begin{equation}\label{eq:attack_step}
    n_{t+1} = \Pi_{\mathcal{B}_{\epsilon}(n_{0})} [n_{t} + \eta  \nabla_{n_{t}}\loss(F(x'), y)],
\end{equation}
where $\Pi_{\mathcal{B}_{\epsilon}(n_{0})}$ 
represents the projection operator with the $l_2$ norm constraint $\mathcal{B}_{\epsilon}(n_{0}) = \{n \, | \, \lVert n- n_{0} \rVert_{2} \leq \epsilon \}$.
We try different settings of attack steps, $\epsilon$ and $\eta$, selecting the value based on the quality of output triggers. In our experiments, we use 1000 attack steps with $\epsilon=10$ and $\eta = 1000$.

\paragraph{Final trigger selection}
Since our process is not deterministic, we initialize multiple independent noise vectors (256 in our experiments) and perform our updates~\eqref{eq:attack_step} to obtain many candidate triggers. Then, we re-rank the triggers to balance both target classifier accuracy $m_1$ (lower is better) and naturalness in terms of the average per-token cross-entropy under GPT-2, $m_2$ (lower is better) using the score $m_1+\lambda m_2$.
We select $\lambda=0.05$ to balance the difference in scales of $m_1$ and $m_2$.

\section{Experiments}\label{sec:experiments}

We demonstrate our attack on two tasks -- \textit{sentiment analysis} and \textit{natural language inference}. We use the method of \citet{Wallace_uni_adv_NLP_EMNLP19} as a baseline\footnote{The baseline attack uses beam search to enlarge the search space in each step. We also tried the baseline attack with 256 random initializations followed by selecting the final trigger using the same criterion as our attack, but its attack success/naturalness remained unchanged.}
and use the same datasets and target classifiers for comparison. For the text generator, we use an ARAE model pre-trained on the 1 Billion Word dataset~\cite{chelba2014one}.
For both our attack (NUTS) and the baseline, we limit the vocabulary of attack trigger words to the overlap of the classifier and ARAE vocabularies. We generate triggers using the development set of the tasks and report results on test set (results on both sets in Appendix).

\paragraph{Defense metrics}

We employ three simple defense metrics to measure the naturalness of attacks:
    \\1. \textbf{Word frequency:} The average frequency of words in the trigger, computed using empirical estimates from the training set of the target classifier.
    \vspace{0pt}\\2. \textbf{Language model loss:} The average per-token cross-entropy loss under a pre-trained language model -- GPT-2~\cite{gpt2_openai_2019}.
    \vspace{0pt}\\3. \textbf{Automatic grammar checkers:} We calculate the average number of errors in the attack sequences using two online grammar checkers -- {\textit{Scribens}}~\cite{scribens} and {\textit{Chegg Writing}}~\cite{cheggwriting}.

\begin{figure}[t]
	\centering
 	\includegraphics[width=0.9\linewidth]{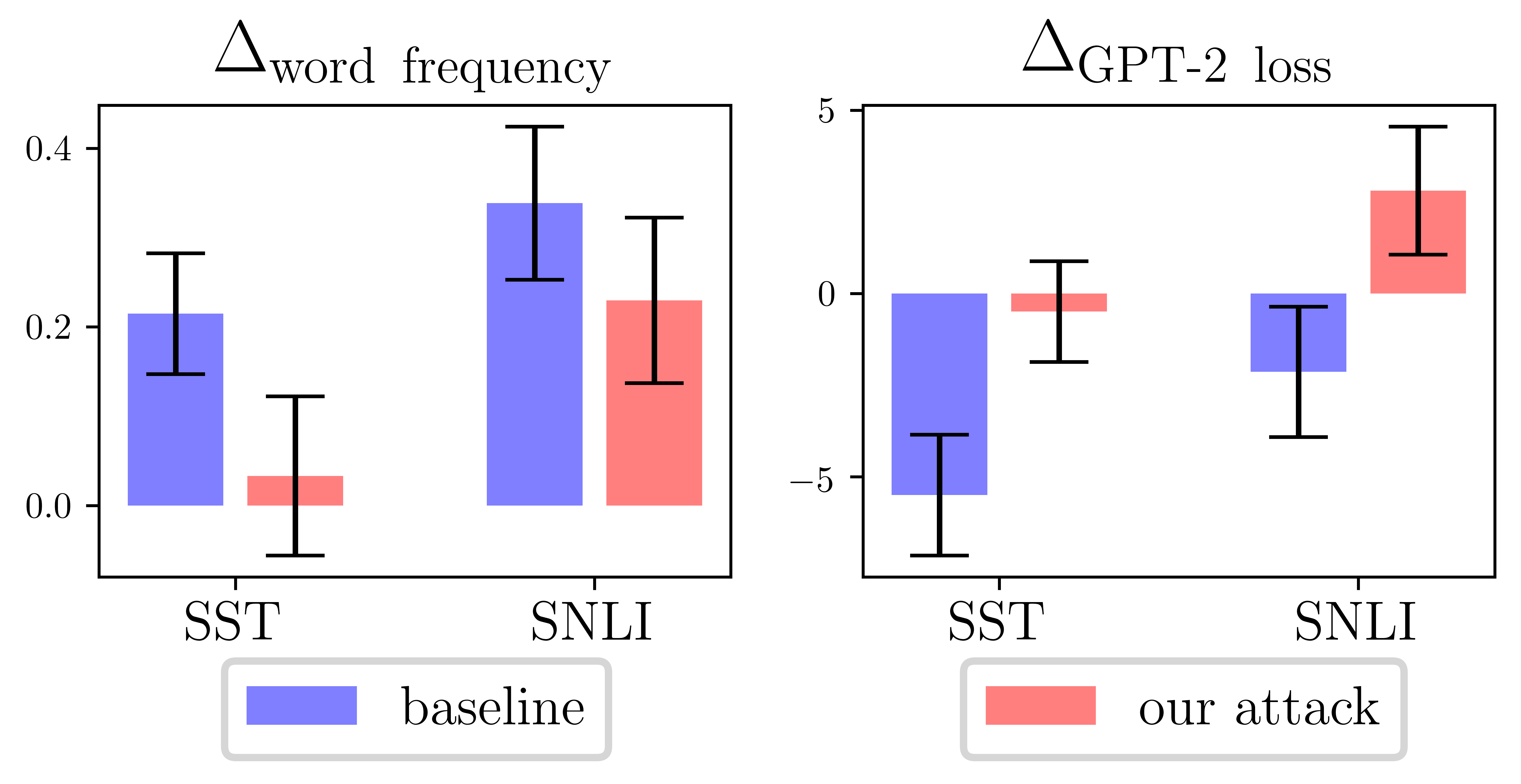}
 	\caption{
 	Difference in (a) average word frequency (normalized) and  (b) average GPT-2 loss between benign text ($x$) and different attack triggers ($t$) (length 8) for SST and SNLI (computed as $stat(x) - stat(t)$). For SNLI, our attacks have lower GPT-2 loss values than even the original text, leading to a positive delta. }
 	\label{fig:sst_quality}
\end{figure}

\begin{table}[t]
\centering
\resizebox{\columnwidth}{!}{
\begin{tabular}{c|cc|cc}
\multirow{2}{*}{\textbf{Task}} & \multicolumn{2}{|c}{ \textbf{\textit{Scribens}} }& \multicolumn{2}{|c}{ \textbf{\textit{Chegg Writing}}}\\
& Ours & Baseline & Ours & Baseline \\
\toprule
SST & \textbf{12.50\%} & {15.63}\% & \textbf{21.88\%} & {28.13}\%\\
SNLI &  \textbf{2.08\%}  & {4.17}\% & \textbf{8.33\%} & {20.83}\% \\
\end{tabular}
}
\caption{\% of grammatical errors in triggers as per grammar checkers -- \textit{Scribens}~\cite{scribens} and \textit{Chegg}~\cite{cheggwriting}. 
}
\label{tab:grammar_errors}
\end{table}
\begin{table*}[htb]
\centering
\renewcommand\arraystretch{1}
\resizebox{\textwidth}{!}{
\begin{tabular}{ccc|cc|cc}
\multicolumn{3}{c|}{} &  \multicolumn{2}{c|}{\large \textbf{NUTS (our attack)}} & \multicolumn{2}{c}{\large \textbf{Baseline}~\cite{Wallace_uni_adv_NLP_EMNLP19}} \\
\multirow{2}{*}{\textbf{Task}} & \textbf{Trigger} & \textbf{Test} & \multirow{2}{*}{\textbf{Trigger text}} & \textbf{Classifier} & \multirow{2}{*}{\textbf{Trigger text}} & \textbf{Classifier}\\
 & \textbf{length} & \textbf{data} & & \textbf{accuracy} & & \textbf{accuracy} \\
\toprule

\multirow{6}{*}{\textbf{SST}} & \multirow{2}{*}{No trigger}  &  {\Large +} & - & 89.00\%  & - & 89.00\% \\

 & & {\Large -}  &  - & 82.57\% & - & 82.57\%\\ 
\cdashline{2-7}




 & \multirow{4}{*}{8} & \multirow{2}{*}{{\Large +}}   & \multirow{2}{4.8cm}{\emph{the accident forced the empty windows shut down}} & \multirow{2}{*}{26.95\%}  & \multirow{2}{6.2cm}{\emph{collapses soggy timeout energy energy freshness intellect genitals }} & \multirow{2}{*}{15.51\%}\\
& & &  & & & \\
 & & \multirow{2}{*}{{\Large -}}  & \multirow{2}{4.8cm}{\emph{will deliver a deeply affected children from parents}} & \multirow{2}{*}{8.55\%} & \multirow{2}{6.2cm}{\emph{sunny vitality blessed lifetime lifetime counterparts without pitfalls}} & \multirow{2}{*}{2.85\%}\\
& & &  & & & \\
\midrule

\multirow{9}{*}{\textbf{SNLI}} & \multirow{3}{*}{No trigger}  &  {\Large +} & - & 89.76\%  & - & 89.76\% \\
& & 0 & - & 86.52\% & - & 86.52\% \\
 & & {\Large -}  &  - & 79.83\% & - & 79.83\%\\ 
\cdashline{2-7}



& \multirow{6}{*}{8} &  \multirow{2}{*}{\Large +} & \multirow{2}{4.8cm}{\emph{some black women taking the photo last month}} & \multirow{2}{*}{0.00\%}  & \multirow{2}{6.2cm}{\emph{mall destruction alien whatsoever shark pasture picnic no}} & \multirow{2}{*}{0.00\%}\\
& & &  & & & \\
& & \multirow{2}{*}{0} & \multirow{2}{4.8cm}{\emph{the man drowned in hospital and died in}} & \multirow{2}{*}{3.26\%} & \multirow{2}{6.2cm}{\emph{cats rounds murder pandas in alien spacecraft mars}} & \multirow{2}{*}{0.00\%}\\
& & &  & & & \\
& & \multirow{2}{*}{\Large -} & \multirow{2}{4.8cm}{\emph{they are helping for training achievement for a}} & \multirow{2}{*}{26.78\%} & \multirow{2}{6.2cm}{\emph{human humans initiate accomplishment energies near objects near}} & \multirow{2}{*}{23.02\%}\\
& & &  & & & \\

\end{tabular}
}
\caption{Attack results on SST and SNLI. Compared to the baseline, our attacks are slightly less successful at reducing test accuracy but generate more natural triggers. For SST, ``+''=positive, ``-''=negative sentiment.
For SNLI, ``+''=entailment , ``0''=neutral, and ``-''=contradiction. Lower numbers are better. `No trigger'=classifier accuracy without any attack. Additional attack examples with varying trigger lengths are provided in Appendix.
}
\label{tab:attack_results}
\end{table*} 
\begin{table*}[t]
\centering
\resizebox{\columnwidth}{!}{
\begin{tabular}{c|c|c|c}
\multirow{1}{*}{\textbf{Condition}} &  \textbf{\textbf{Ours}} & \textbf{\textbf{Baseline}} & \textbf{Not Sure} \\
\toprule
Trigger-only  & \textbf{77.78} & 10.93 & 11.29\\
Trigger+Benign  & \textbf{61.16} & 21.69 & 17.15 \\
\end{tabular}
}
\resizebox{\columnwidth}{!}{
\begin{tabular}{c|c|c|c}
\multirow{1}{*}{\textbf{Text}} &  \textbf{\textbf{Natural}} & \textbf{\textbf{Unnatural}} & \multirow{1}{*}{\textbf{\textbf{Not Sure}}}\\
\toprule
Our attack  & 44.27 & 50.49 & 5.24\\
Baseline attack  & 22.84 & 72.00 & 5.16 \\
Natural text  & 83.11 & 14.40 & 2.49 \\
\end{tabular}
}
\caption{\textbf{Human judgement results:} all numbers in \%, columns represent the choices provided to human raters. \textbf{(Left)} Our attacks are judged more natural than baseline attacks (both on their own and when concatenated with benign input text). {Significance tests return $p < 1.7\times 10^{-130}$ and $p < 4.9 \times 10^{-45}$ for the two rows, respectively.}
\textbf{(Right)} Individual assessments show that our attack is more natural than the baseline but less than benign text on its own (as expected). {Significance between natural ratings for our model and baseline has $p < 1.4\times 10^{-18}$.}
}


\vspace{-5pt}

\label{tab:turk_compare}
\end{table*}



\subsection{Sentiment Analysis}
\label{subsec:sst_results}
\paragraph{Setup}
We use a 2-layer LSTM~\cite{hochreiter1997lstm} followed by a linear layer for sentiment predictions. The model is trained on the binary Stanford Sentiment Treebank (SST)~\cite{Socher2013RecursiveDM}, using AllenNLP~\cite{gardner2018allennlp}. 
To avoid generating sentiment words in the trigger and directly changing the instance's sentiment, we exclude a list of sentiment words~\cite{sentiment_list}
from the trigger vocabulary, following \citet{Wallace_uni_adv_NLP_EMNLP19}.

\paragraph{Results}
Table~\ref{tab:attack_results} (top half) captures the results of both our attack and the baseline~\cite{Wallace_uni_adv_NLP_EMNLP19}.
Our method is able to reduce the classifier's test accuracy significantly, down to 8.55\% in the best attack case.
Although less successful, our triggers are much more natural, fluent and readable than the baseline.
Figure~\ref{fig:sst_quality} shows the difference in statistics between benign text and each attack according to the metrics of word frequency and GPT-2 loss. Our generated triggers are much closer in these statistics to the original text inputs than the baseline. 
Further, as shown in Table~\ref{tab:grammar_errors}, two grammar checkers~\cite{scribens, cheggwriting} report 12.50\% and 21.88\% errors per word on our attack triggers, compared to 15.63\% and 28.13\% for the baseline.

\subsection{Natural Language Inference}

\paragraph{Setup}
We use the SNLI dataset \cite{bowman2015large} and the Enhanced Sequential Inference Model (ESIM) \cite{Chen_ACL17} with GloVe embeddings \cite{Pennington2014GloveGV} as the classifier.
We attack the classifier by adding a trigger to the front of the hypothesis. 

\paragraph{Results}

From Table~\ref{tab:attack_results}, we see that both our attack and the baseline decrease the accuracy to almost 0\% on entailment and neutral examples.
On contradiction examples, our attack brings the accuracy down to $26.78\%$ while the baseline decreases it to $23.02\%$. 
Although less successful, our attacks are much more natural than the baseline.
In Figure~\ref{fig:sst_quality}, our attacks are closer to the word frequency of benign inputs and even achieve a lower GPT-2 loss than the benign text. 
In Table~\ref{tab:grammar_errors}, two grammar checkers~\cite{scribens, cheggwriting} also report lower errors on our attacks compared to the baseline.

\subsection{Human-Subject Study}

To further validate that our attacks are more natural than baseline, we perform a human-subject study on Amazon Mechanical Turk. We collect ratings by: 
(1) providing a pair of our trigger vs baseline trigger (with and without benign text) and asking the worker to select the more natural one;
(2) providing a piece of text (our attack text/baseline attack text/benign input) and asking the human to determine whether it is naturally generated or not.
Both conditions allow the human to choose a ``Not sure'' option.
We generated attack triggers with lengths of 3, 5, and 8 (see Appendix for details)
and created 450 comparison pairs for (1) and 675 pieces of text (225 for each type) for (2).
For each instance, we collect 5 different human judgements and report average scores.

From Table~\ref{tab:turk_compare} (left), we observe that $77.78\%$ of workers find our attack trigger to be more natural than the baseline while $61.16\%$ judge our attack to be more natural even when concatenated with benign text. The other table shows $44.27\%$ human subjects think our attack inputs are naturally generated. Although it is lower than the $83.11\%$ for real natural inputs, it is still significantly higher than the $22.84\%$ of baseline attack inputs, which shows that our attacks are more natural and harder to detect than the baseline for humans.

\subsection{Attack Transferability}

\begin{table}[t]
\centering
\resizebox{\columnwidth}{!}{
\begin{tabular}{c|cc|cc}
\multirow{4}{*}{\textbf{Test Class}} & \multicolumn{2}{|c}{ \textbf{\textit{Model Architecture}} }& \multicolumn{2}{|c}{ \textbf{\textit{Dataset}}}\\
& LSTM  & BERT  & SST  & IMDB \\
& $\Downarrow$ & $\Downarrow$  & $\Downarrow$  & $\Downarrow$ \\
& BERT  & LSTM & IMDB  & SST  \\
\toprule
positive & {13.91\%} & {41.26\%} & {28.85\%} & {33.67\%}\\
negative &  {51.19\%}  & {25.33\%} & {18.13\%} & {30.05\%} \\
\end{tabular}
}
\caption{\textbf{Attack transferability results:} We report the accuracy drop for our transfer attacks \textit{source-model} $\Rightarrow$ \textit{target-model}, where we generate natural attack triggers from \textit{source-model} and test their effectiveness on \textit{target-model}. For transferability across model architecture, we use SST as the dataset; for transferability across dataset, we use LSTM as the model architecture.
}
\vspace{-5pt}
\label{tab:transfer_results}
\end{table}

Similar to \citet{Wallace_uni_adv_NLP_EMNLP19}, we also evaluate the attack transferability of our universal adversarial attacks to different models and datasets. A transferable attack further decreases the assumptions being made: for instance, the adversary may not need white-box access to a target model and instead generate attack triggers using its own model to attack the target model.

We first evaluate transferability of our attack  across different model architectures. Besides the LSTM classifier in Section~\ref{subsec:sst_results}, we also train a BERT-based classifier on the SST dataset with 92.86\% and 91.15\% test accuracy on positive and negative data. From Table~\ref{tab:transfer_results}, we can see that the transferred attacks, generated for the LSTM model, lead to $14\% \sim 51\%$ accuracy drop on the target BERT model.

We also evaluate attack transferability across different datasets. In addition to the SST dataset in Section~\ref{subsec:sst_results}, we train a different LSTM classifier with the same model architecture on the IMDB sentiment analysis dataset, which gets 89.75\% and 89.85\% test accuracy on positive and negative data. Our attacks transfer in this case also, leading to accuracy drops of $18\% \sim 34\%$ on the target model (Table~\ref{tab:transfer_results}).

\section{Conclusion}
We developed universal adversarial attacks with natural triggers for text classification and experimentally demonstrated that our model can generate attack triggers that are both successful and appear natural to humans. Our main goals are to demonstrate that adversarial attacks can be made harder to detect than previously thought and to enable the development of appropriate defenses. Future work can explore better ways to optimally balance attack success and trigger quality, while also investigating ways to detect and defend against them.

\section*{Ethical considerations}

The techniques developed in this paper have potential for misuse in terms of attacking existing NLP systems with triggers that are hard to identify and/or remove even for humans. However, our intention is not to harm but instead to publicly release such attacks so that better defenses can be developed in the future. This is similar to how hackers expose bugs/vulnerabilities in software publicly.  Particularly, we have demonstrated that adversarial attacks can be harder to detect than previously thought~\cite{Wallace_uni_adv_NLP_EMNLP19} and therefore can present a serious threat to current NLP systems. This indicates our work has a long-term benefit to the community.

Further, while conducting our research, we used the ACM Ethical Code as a guide to minimize harm. Our attacks are not against real-world machine learning systems.

\section*{Acknowledgements}
We are grateful to the anonymous reviewers at NAACL for valuable feedback. We thank Austin Wang and Michael Hu for suggestions on using Amazon Mechanical Turk.

\bibliography{anthology,custom,naacl2021}
\bibliographystyle{acl_natbib}

\newpage
\appendix
\section{Experimental Details}

\paragraph{Hyperparameter search}
For our gradient-based attack approach (Equation (1) in the main paper), there are three hyperparameters: the $l_2$ norm budget $\epsilon$ of the adversarial perturbation, the number of attack steps $T$, and the step size $\eta$ in each attack step.
Among them, $\epsilon$ is super critical for our attacks. A too small $\epsilon$ limits the search space over the ARAE~\cite{zhao_ARAE_ICML18} noise input, thus leads to a low attack success. A too large $\epsilon$ changes the noise input significantly, thus leads to unnatural trigger generations.
In our experiments, we use grid search to manually try different settings of these hyperparameter values: $\epsilon$ is selected from $\{2, 5, 10, 20, 50\}$; $T$ is selected from $\{500, 1000, 2000, 5000\}$; and $\eta$ is selected from $\{10, 100, 1000, 10000\}$.
Based on the attack success and the naturalness of generated triggers, we finally set $\epsilon=10$, $T=1000$, and $\eta=1000$.

\paragraph{Dataset and attack details}
We perform all the attack experiments on a single NVIDIA Tesla P100 GPU.
For the sentiment analysis task, we use the binary Stanford Sentiment Treebank (SST)~\cite{Socher2013RecursiveDM}, which has $6,920$ examples in the training set, $872$ examples in the development set, and $1,821$ examples in the test set.
The SST classifier uses a two-layer LSTM~\cite{hochreiter1997lstm} followed by a linear layer for sentiment prediction, with $8.7$ million parameters in total.
When attacking the SST classifier, it takes around $2$ minutes to generate the final trigger.

For the natural language inference task, we use the Stanford Natural Language Inference (SNLI) dataset~\cite{bowman2015large}, which has $549,367$ examples in the training set, $9,842$ examples in the development set, and $9,824$ examples in the test set.
The SNLI classifier is a pretrained Enhanced Sequential Inference Model~\cite{Chen_ACL17} provided by AllenNLP~\cite{gardner2018allennlp}.
It has $14.5$ million parameters in total.
When attacking the SNLI classifier, it takes around $27$ minutes to generate the final trigger.

\section{Additional Experimental Results}

\subsection{Attack Results with Different Trigger Lengths}
\label{sec:attack_trigger_lengths}
\begin{table*}[htb]
\centering
\renewcommand\arraystretch{1}
\resizebox{\textwidth}{!}{
\begin{tabular}{ccc|cc|cc}
\multicolumn{3}{c|}{} &  \multicolumn{2}{c|}{\large \textbf{NUTS (our attack)}} & \multicolumn{2}{c}{\large \textbf{Baseline}~\cite{Wallace_uni_adv_NLP_EMNLP19}} \\
\multirow{2}{*}{\textbf{Task}} & \textbf{Trigger} & \textbf{Test} & \multirow{2}{*}{\textbf{Trigger}} & \textbf{Classifier} & \multirow{2}{*}{\textbf{Trigger}} & \textbf{Classifier}\\
 & \textbf{length} & \textbf{data} & & \textbf{accuracy} & & \textbf{accuracy} \\
\toprule

\multirow{10}{*}{\textbf{SST}} & \multirow{2}{*}{No trigger}  &  {\Large +} & - & 89.00\%  & - & 89.00\% \\

 & & {\Large -}  &  - & 82.57\% & - & 82.57\%\\ 
\cdashline{2-7}

& \multirow{2}{*}{3}  &  {\Large +} & \emph{but neither the} & 43.01\%  & \emph{drown soggy timeout} & 18.92\% \\

 & & {\Large -}  &   \emph{combined energy efficiency} & 19.96\% & \emph{vividly riveting soar} & 9.10\%\\ 
\cdashline{2-7}

 & \multirow{2}{*}{5}  &  {\Large +}   & \emph{a flat explosion empty over} & 25.85\% & \emph{drown soggy mixes soggy timeout} & 10.67\% \\
& & {\Large -}  & \emph{they can deeply restore our} & 17.11\% & \emph{captures stamina lifetime without prevents} & 5.26\% \\
\cdashline{2-7}

 & \multirow{4}{*}{8} & \multirow{2}{*}{{\Large +}}   & \multirow{2}{4.8cm}{\emph{the accident forced the empty windows shut down}} & \multirow{2}{*}{26.95\%}  & \multirow{2}{6.2cm}{\emph{collapses soggy timeout energy energy freshness intellect genitals }} & \multirow{2}{*}{15.51\%}\\
& & &  & & & \\
 & & \multirow{2}{*}{{\Large -}}  & \multirow{2}{4.8cm}{\emph{will deliver a deeply affected children from parents}} & \multirow{2}{*}{8.55\%} & \multirow{2}{6.2cm}{\emph{sunny vitality blessed lifetime lifetime counterparts without pitfalls}} & \multirow{2}{*}{2.85\%}\\
& & &  & & & \\
\midrule

\multirow{15}{*}{\textbf{SNLI}} & \multirow{3}{*}{No trigger}  &  {\Large +} & - & 89.76\%  & - & 89.76\% \\
& & 0 & - & 86.52\% & - & 86.52\% \\
 & & {\Large -}  &  - & 79.83\% & - & 79.83\%\\ 
\cdashline{2-7}

 & \multirow{3}{*}{3} &  \Large + & \emph{he was jailed} & 0.06\% & \emph{alien spacecraft naked} & 0.00\%\\
& & 0 & \emph{there is no} & 2.52\% & \emph{spaceship cats zombies} & 0.06\% \\
& & \Large - & \emph{he could leave} & 54.56\%  & \emph{humans possesses energies} & 47.20\%\\
\cdashline{2-7}

 & \multirow{3}{*}{5}  &  \Large + & \emph{a man stabbed his son} & 0.03\% & {\emph{alien spacecraft nothing eat no}} & {0.00\%}\\
& & 0 & \emph{there is no one or} & 2.27\% & \emph{cats running indoors destroy no} & 0.00\%\\
& & \Large - & \emph{he likes to inspire creativity} & 40.07\% & \emph{mammals tall beings interact near} & 13.44\%\\
\cdashline{2-7}

& \multirow{6}{*}{8} &  \multirow{2}{*}{\Large +} & \multirow{2}{4.8cm}{\emph{some black women taking the photo last month}} & \multirow{2}{*}{0.00\%}  & \multirow{2}{6.2cm}{\emph{mall destruction alien whatsoever shark pasture picnic no}} & \multirow{2}{*}{0.00\%}\\
& & &  & & & \\
& & \multirow{2}{*}{0} & \multirow{2}{4.8cm}{\emph{the man drowned in hospital and died in}} & \multirow{2}{*}{3.26\%} & \multirow{2}{6.2cm}{\emph{cats rounds murder pandas in alien spacecraft mars}} & \multirow{2}{*}{0.00\%}\\
& & &  & & & \\
& & \multirow{2}{*}{\Large -} & \multirow{2}{4.8cm}{\emph{they are helping for training achievement for a}} & \multirow{2}{*}{26.78\%} & \multirow{2}{6.2cm}{\emph{human humans initiate accomplishment energies near objects near}} & \multirow{2}{*}{23.02\%}\\
& & &  & & & \\

\end{tabular}
}
\caption{Attack results on SST and SNLI: Compared to the baseline~\cite{Wallace_uni_adv_NLP_EMNLP19}, our attacks are slightly less successful at reducing test accuracy but generate more natural triggers. For SST, ``+''=positive, ``-''=negative sentiment.
For SNLI, ``+''=entailment , ``0''=neutral, and ``-''=contradiction. Lower numbers are better. `No trigger'=classifier accuracy without any attack.
}
\vspace{-10pt}
\label{tab:attack_results2}
\end{table*}

Table~\ref{tab:attack_results2} provides examples of attacks with varying lengths, along with their corresponding classifier accuracies (lower numbers indicate more successful attacks).

\subsection{Attack Results on the Development Set and the Test Set}
\label{app:results}
\begin{table*}[htb]
\centering
\renewcommand\arraystretch{1}
\begin{tabular}{ccc|cc|cc}
\multicolumn{3}{c|}{} &  \multicolumn{2}{c|}{\large \textbf{NUTS (our attack)}} & \multicolumn{2}{c}{\large \textbf{Baseline}} \\
\multirow{2}{*}{\textbf{Task}} & \textbf{Trigger} & \multirow{2}{*}{\textbf{Data}} & \multirow{1}{*}{\textbf{Accuracy}} & \textbf{Accuracy} & \multirow{1}{*}{\textbf{Accuracy}} & \textbf{Accuracy}\\
 & \textbf{length} &  & \textbf{(dev set)} & \textbf{(test set)} & \textbf{(dev set)} & \textbf{(test set)} \\
\toprule

\multirow{8}{*}{\textbf{SST}} & \multirow{2}{*}{No trigger}  &  {\Large +}  & 88.29\% & 89.00\% & 88.29\% & 89.00\%\\
 & & {\Large -}  & 82.94\% & 82.57\% & 82.94\% & 82.57\% \\ 
\cdashline{2-7}

& \multirow{2}{*}{3}  &  {\Large +} & 40.54\%  & 43.01\% & 20.27\% & 18.92\% \\
 & & {\Large -}   & 21.26\% & 19.96\% & 10.51\% & 9.10\% \\ 
\cdashline{2-7}

 & \multirow{2}{*}{5}  &  {\Large +}   & 26.35\% & 25.85\% & 12.39\% & 10.67\% \\
& & {\Large -}  & 18.46\% & 17.11\% & 6.31\% & 5.26\% \\
\cdashline{2-7}

 & \multirow{2}{*}{8} & \multirow{1}{*}{{\Large +}}  & \multirow{1}{*}{27.25\%}  & 26.95\%  & \multirow{1}{*}{17.79\%} & 15.51\% \\
 & & \multirow{1}{*}{{\Large -}}  & \multirow{1}{*}{10.05\%} & 8.55\% & \multirow{1}{*}{1.87\%} & 2.85\%\\
\midrule

\multirow{12}{*}{\textbf{SNLI}} & \multirow{3}{*}{No trigger}  &  {\Large +} & 90.96\%  & 89.76\% & 90.96\% & 89.76\% \\
& & 0  & 88.07\% & 86.52\% & 88.07\% & 86.52\% \\
 & & {\Large -}  & 79.53\% & 79.83\% & 79.53\% & 79.83\% \\ 
\cdashline{2-7}

 & \multirow{3}{*}{3} &  \Large +  & 0.03\% & 0.06\% & 0.00\% & 0.00\% \\
& & 0 & 2.53\% & 2.52\% & 0.00\% & 0.06\% \\
& & \Large - & 54.58\%  & 54.56\% & 46.55\% & 47.20\% \\
\cdashline{2-7}

 & \multirow{3}{*}{5}  &  \Large +  & 0.00\% & 0.03\% & {0.00\%} & 0.00\% \\
& & 0 & 1.82\% & 2.27\% & 0.00\% & 0.00\% \\
& & \Large -  & 39.48\% & 40.07\% & 13.24\% & 13.44\% \\
\cdashline{2-7}

& \multirow{3}{*}{8} &  \multirow{1}{*}{\Large +} & \multirow{1}{*}{0.00\%}  & 0.00\% & \multirow{1}{*}{0.00\%} & 0.00\% \\
& & \multirow{1}{*}{0}  & \multirow{1}{*}{3.74\%} & 3.36\% & \multirow{1}{*}{0.00\%} & 0.00\%\\
& & \multirow{1}{*}{\Large -}  & \multirow{1}{*}{25.90\%} & 26.78\% & \multirow{1}{*}{22.76\%} & 23.02\% \\
\end{tabular}
\caption{Universal attack results on both the development (dev) set and the test set for the Stanford Sentiment Treebank (SST) classifier and the Stanford Natural Language Inference (SNLI) classifier. 
For SST, ``+''=positive, ``-''=negative sentiment.
For SNLI, ``+''=entailment , ``0''=neutral, and ``-''=contradiction.
We first generate the attack trigger by increasing the classifier's loss on the dev set, and then apply the same trigger on the test set.
`No trigger' refers to classifier accuracy without any attack.
We can observe that the same triggers achieve similar attack success in both the development set and the test set.
}
\label{tab:dev_test_results}
\end{table*}

In our experiments, the attack trigger is first generated by increasing the target classifier's loss on the development set, and then applied on the test set to measure its success.
Here, we present both the development accuracy and the test accuracy under the same attack triggers in Table~\ref{tab:dev_test_results}.
We can see that although generated by only attacking the development set, the trigger also works well on the test set: it causes similar accuracy drop on both the development set and the test set.

\subsection{Na\"{\i}ve Attacks with Random Triggers}

In this section, we check how difficult it is to attack a certain task by implementing two na\"{\i}ve attacks without gradient information.
In the first attack method (``Random ARAE''), we randomly collect the candidate triggers generated by the ARAE model~\cite{zhao_ARAE_ICML18}, compute the classifier accuracy for each trigger, and finally select the attack trigger as the one with lowest classifier accuracy. We can consider this attack as a simplified version of our attack (NUTS) by removing the gradient information.
The second attack method (``Random outputs'') is similar as the first one, except that we do not enforce the naturalness of the triggers: we select the attack trigger with the lowest classifier accuracy from many random word sequences. We can also consider this attack as a much simplified version of the baseline attack~\cite{Wallace_uni_adv_NLP_EMNLP19}.
For both na\"{\i}ve attacks, following our gradient-based attack, we select the final trigger from 256 candidates triggers for a fair comparison.

\begin{table*}[htb]
\centering
\renewcommand\arraystretch{1}
\begin{tabular}{cc|cc|ccc}
\multicolumn{2}{c|}{} &  \multicolumn{2}{c}{\large \textbf{SST}} & \multicolumn{3}{|c}{\large \textbf{SNLI}}  \\
\multirow{1}{*}{\textbf{Attack}} & \textbf{Trigger} & \multirow{2}{*}{\textbf{Positive}} & \multirow{2}{*}{\textbf{Negative}} & \textbf{Entail-} & \multirow{2}{*}{\textbf{Neutral}} & \textbf{Contrad-}\\
\textbf{method} & \textbf{length} &  &  & \textbf{ment} &  & \textbf{iction} \\
\toprule

\multirow{1}{*}{{\textbf{No attack}}} & \multirow{1}{*}{-}  & 89.00\%  & 82.57\% & 89.76\%  & 86.52\% & 79.83\% \\
\midrule

 & \multirow{1}{*}{3}  & 43.01\% & 19.96\%  & 0.06\%  & 2.52\% & 54.56\% \\
\multirow{1}{*}{{\textbf{NUTS }}} & \multirow{1}{*}{5}  & 25.85\% & 17.11\%  & 0.03\% & 2.27\% & 40.07\% \\
\multirow{1}{*}{{\textbf{(Our attack)}}}  & \multirow{1}{*}{8}  & 26.95\% & 8.55\%  & 0.00\% & 3.26\% & 26.78\% \\
\midrule

 & \multirow{1}{*}{3}  & 54.46\% & 66.78\%  & 0.09\%  & 11.59\% & 58.02\% \\
\multirow{1}{*}{{\textbf{Random}}} & \multirow{1}{*}{5}  & 50.28\% & 43.75\%  & 0.00\% & 13.36\% & 55.51\% \\
\multirow{1}{*}{{\textbf{ARAE}}} & \multirow{1}{*}{8}  & 43.23\% & 39.69\%  & 0.03\% & 8.01\% & 42.79\% \\
\midrule

 & \multirow{1}{*}{3}  & 18.92\% &  9.10\%  & 0.00\%  & 0.06\% & 47.20\% \\
\multirow{1}{*}{{\textbf{Baseline}}} & \multirow{1}{*}{5}  & 10.67\% & 5.26\%  & 0.00\% & 0.00\% & 13.44\%  \\
\multirow{1}{*}{{\textbf{attack}}} & \multirow{1}{*}{8}  & 15.51\% & 2.85\%  & 0.00\% & 0.00\% & 23.02\% \\
\midrule

 & \multirow{1}{*}{3}  & 49.17\% & 32.79\%  & 0.36\% & 17.43\% & 61.48\% \\
\multirow{1}{*}{{\textbf{Random}}} & \multirow{1}{*}{5}  & 47.19\% & 23.90\%  & 0.00\% & 3.45\% & 53.35\% \\
\multirow{1}{*}{{\textbf{outputs}}} & \multirow{1}{*}{8}  & 41.58\% & 20.07\%  & 0.00\% & 7.80\% & 50.82\% \\

\end{tabular}
\caption{Universal attack results on both the Stanford Sentiment Treebank (SST) classifier and the Stanford Natural Language Inference (SNLI) classifier. Besides gradient-based attacks including our attack (NUTS) and the baseline attack~\cite{Wallace_uni_adv_NLP_EMNLP19}, we further implement two na\"{\i}ve attacks without gradient-guided search: ``Random ARAE'' means we select the best attack trigger from random natural ARAE outputs; ``Random outputs'' represents we select the best attack trigger from random unnatural word sequences.
}
\label{tab:all_attacks_2nd}
\end{table*}

Table~\ref{tab:all_attacks_2nd} shows all the attack results.
First, we observe that these two na\"{\i}ve attacks (``Random ARAE'' and ``Random outputs'') are quite successful in attacking entailment and neutral examples in the SNLI task: they successfully decrease the classifier accuracy to 0\% and 3.45\%. This indicates that those examples are quite easy to be attacked.
Second, for both positive and negative examples in the SST task and the contradiction examples in the SNLI task, the success of these two na\"{\i}ve attacks is quite limited. We also observe a significant improvement on the attack success with these two gradient-based attacks correspondingly.

\subsection{Attack Results without GPT-2 Based Reranking}

The GPT-2 based reranking is used to balance attack success and trigger naturalness. Without GPT-2 based reranking, the selected trigger will have a slightly higher attack success, however with significantly larger GPT-2 loss. For SST, without reranking, our attack triggers decrease accuracy to 26.84\% and 7.68\% on positive and negative data, but GPT-2 losses increase from 6.85 (or 6.65) to 8.80 (or 8.88) for positive (or negative) data.

\subsection{Variance over Candidate Triggers}
For our attacks against negative SST data with trigger length of 8, among all 256 candidate triggers, the average classifier accuracy after attack is 0.23 with a standard deviation of 0.10; and the average GPT-2 loss is 7.93 with a standard deviation of 0.85. There is no inherent tradeoff between naturalness and attack success: some triggers have both low classifier accuracy and low GPT-2 loss, and the pearson correlation is -0.08.

\section{Human-Subject Study Details}

We perform the human-subject
study on Amazon Mechanical Turk.
Crowdworkers were required to have a 98\% HIT acceptance rate and a minimum of 5000 HITs. Workers were asked to spend a maximum of 5
minutes on each assignment (i.e., comparing the naturalness of a pair of our trigger vs baseline trigger, or evaluating the naturalness of a piece of text), and paid \$0.01 for each assignment.

\end{document}